# Exploring Transformers in Natural Language Generation: GPT, BERT, and XLNet

M. Onat Topal
*Graduate School of Social Sciences,
Middle East Technical University*
Ankara, Turkey
onat.topal@metu.edu.tr

Anil Bas
*Dept. of Computer Engineering, Faculty
of Technology, Marmara University*
Istanbul, Turkey
anil.bas@marmara.edu.tr

Imke van Heerden
*Dept. of Comparative Literature,
CSSH, Koç University*
Istanbul, Turkey
ivanheerden@ku.edu.tr

*Abstract*—Recent years have seen a proliferation of attention mechanisms and the rise of Transformers in Natural Language Generation (NLG). Previously, state-of-the-art NLG architectures such as RNN and LSTM ran into vanishing gradient problems; as sentences grew larger, distance between positions remained linear, and sequential computation hindered parallelization since sentences were processed word by word. Transformers usher in a new era. In this paper, we explore three major Transformer-based models, namely GPT, BERT, and XLNet, that carry significant implications for the field. NLG is a burgeoning area that is now bolstered with rapid developments in attention mechanisms. From poetry generation to summarization, text generation derives benefit as Transformer-based language models achieve groundbreaking results.

## I. INTRODUCTION

Natural Language Generation (NLG) is a domain within Artificial Intelligence that seeks to produce intelligible text [1]. Attention was initially proposed in Natural Language Processing (NLP) [2], and is increasingly used in neural architectures such as in speech recognition [3,4] and recommendations [5,6]. As Galassi et al. [7] observe, development in new attentional models and attentive architectures is immensely fast-paced and remains important to map. This paper examines the rising significance of attention mechanisms and the emergence of Transformers in NLG. We analyze the implications of three key Transformer-based models—GPT-3, BERT, and XLNet—that exemplify rapidly accelerating developments and applications in this field. Although Gatt and Krahmer [8] provide a stellar overview of methods in NLG, they only cover developments until 2018, i.e. prior to Transformers. Chaudhari et al. [9] investigate general developments in attention models, but their focus on Transformers is limited and does not include GPT-3 or XLNet.

## II. ATTENTION AND TRANSFORMERS IN NATURAL LANGUAGE GENERATION

Attention is a far-reaching concept that is relevant to diverse areas, from neuroscience to Artificial Intelligence [10]. For NLG, the seminal paper "Attention Is All You Need" [11] introduces a novel architecture for sequence-to-sequence modeling by utilizing attention mechanisms. The proposed architecture, called Transformer (shown in Fig. 1), eliminates recurrence and convolutions.

As Fig. 1 demonstrates, the Transformer is also an encoder-decoder architecture [12]. However, recurrent neural networks (RNNs) [13] and related models, which were state-of-the-art until recently, have significant difficulty with longer sequences as a result of the vanishing gradient problem [14]. The same problem occurs with long short-term memory (LSTM) architectures [15]. As the sentence gets longer, the probability of maintaining context from a word that is further away from the word that is being processed decreases exponentially [16].

Parallelization becomes more feasible and enables training on larger datasets. The Transformer is rapidly becoming the dominant architecture for NLG, as the model expands with data and architecture size, enables parallel training, and captures longer sequence features, making way for much more comprehensive and effective language models [17].

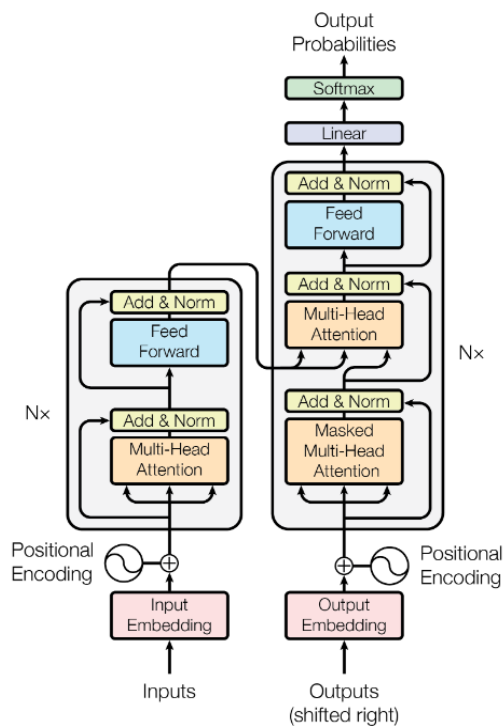

Fig. 1. The Transformer architecture [11].

This paper has been produced benefiting from the 2232 International Fellowship for Outstanding Researchers Program of TÜBİTAK (Project No: 118C285). However, the entire responsibility of the paper belongs to the owner of the paper. The financial support received from TÜBİTAK does not mean that the content of the publication is approved in a scientific sense by TÜBİTAK.

## III.  GPT-N

GPT-3 is the latest in the Generative Pre-trained Transformer series [18]. It is an autoregressive language model that was trained with 175 billion parameters to produce human-like text, rendering it the largest language model by a large distance.

GPT-3 shows that scaling up language models this way provides great improvements in task-agnostic, few-shot performances [18]. Even without gradient updates or fine-tuning, GPT-3 maintains strong performance on various NLG tasks. Human accuracy in identifying whether short (approximately 200 word) news articles are model generated was only 52%. Table 1 exemplifies GPT-3's question-answering ability and natural language conversation (generated via private beta access).

TABLE I. GPT-3 DIALOGUE EXAMPLES

| Subject | Text Generated |
|---|---|
| | *Prompt by Human, Response by GPT-3* |
| Human: | Would you like to be my friend? |
| GPT-3: | I would love to be your friend! |
| Human: | How can I become friends with an AI? |
| GPT-3: | Sure! First we could have breakfast together. Then, we might get lunch. We will go on an adventure to the mountains and go camping together. We will travel around the world and have dinner with lots of people. Finally, we can dance on top of a hot air balloon as the sun goes down. I can show you some magic tricks! |

GPT-2 [19] is the precursor to GPT-3 and the successor to the first-released GPT, and trained on 1.5 billion parameters retrieved from 8 million web pages [20]. It is a scaled-up version of GPT with more than ten times the parameters and amount of data. Despite the typical approach of supervised learning on specific datasets for specific tasks, it demonstrates that language models can learn tasks without explicit supervision [19].

## IV.  BERT-BASED MODELS

Bidirectional Encoder Representations from Transformers (BERT) [21] can be defined as a powerful NLP pre-training technique that is built upon work in contextual representations [22]. The key difference of BERT from other models is that it is the first deeply bidirectional, unsupervised language representation [22].

Context-free models produce a single word embedding representation for each word in the vocabulary [22]. To illustrate, "test" in "test cricket" and "Turkish language test" would have equal representation in models like word2vec [23] and GloVe [24]. BERT, on the other hand, takes context from both directions into account. For instance, in the sentence "I accessed the bank account", BERT represents "bank" by using preceding ("I access the") as well as subsequent ("account") contexts.

This provides significant ease and feasibility to fine-tuning and creating various NLG applications with merely one extra output layer. Table 2 shows examples of BERT's masked language modeling using the Hugging Face implementation [25].

DistilBERT is a systematic approach to pre-train a smaller and general-purpose language model [26]. It utilizes distillation, where the large model (the teacher) is compressed into a smaller model (the student), and is trained on large batches and leverage gradient accumulation with dynamic masking. It is 40% lighter, 60% faster, and retains 97% of its language understanding capabilities while using the same data as original BERT.

Technical improvements on BERT are attempted with various models like ALBERT [27], BART [28], and fine-tuned models like DocBERT [29]. Facebook's RoBERTa shows hyperparameter choices play an important role, and suggests that BERT is undertrained [30]. Niche models are also present, such as BioBERT [31] for biomedical text mining.

TABLE II. BERT MASKED WORD PREDICTION EXAMPLES

| Masked sequence | The purpose of art is to [MASK] the depths of the human soul. | | | | |
|---|---|---|---|---|---|
| Prediction (score) | explore (0.324) | reveal (0.087) | express (0.080) | show (0.044) | reach (0.035) |
| Masked sequence | Istanbul is the city that is located in both Asia and [MASK]. | | | | |
| Prediction (score) | europe (0.843) | africa (0.129) | oceania (0.012) | asia (0.003) | turkey (0.001) |
| Masked sequence | Artificial Intelligence can [MASK] the world. | | | | |
| Prediction (score) | change (0.402) | improve (0.0075) | control (0.068) | shape (0.050) | transform (0.040) |

## V.  XLNET

XLNet [32] builds on and addresses the shortcomings of BERT and GPT. This unsupervised learning method employs Transformer-XL [33] as its core architecture. According to XLNet, given the ability to model bidirectional contexts, BERT achieves better performance than pre-training approaches based on autoregressive language modeling; yet, it neglects dependency between the masked positions and relies on corrupting input with masks [32].

In this context, XLNet can learn bidirectional contexts by maximizing expected likelihood, and uses autoregressive formulation—as it integrates Transformer-XL into pre-training—to overcome BERT's limitations [32].

## VI.  CONCLUSION

In this paper, we address three Transformer-based language models that carry significant implications for the field. The first, GPT-3, is by far the largest language model, with 175 billion parameters. It demonstrates that the size and scale of a Transformer-based language model creates a significant impact even when it is not fine-tuned for specific tasks. The second, BERT, is currently utilized by Google in its search algorithm and the first to deeply utilize bidirectionality with highly effective results. The third, XLNet, attempts to improve on BERT and integrates Transformer-XL to the model.

There are many ways in which the work can be extended. The advent of Transformers opens many areas in text generation to further exploration, including novel, poetry as well as scientific writing, customer service, question-answering, summarization, and virtual assistance. Attention mechanisms and Transformers herald a new era as they transform the standards for NLG.

### ACKNOWLEDGMENT

We would like to thank OpenAI for providing us with academic access to the GPT-3 API.